\documentclass[sigconf,camera-ready]{acmart}

\usepackage{algorithm}

\usepackage{microtype}
\usepackage{subfigure}
\usepackage{booktabs} 
\usepackage{wrapfig}
\usepackage{hyperref}

\usepackage{algpseudocode}

\usepackage{mathtools}

\usepackage[capitalize,noabbrev]{cleveref}

\theoremstyle{plain}
\theoremstyle{definition}
\theoremstyle{remark}

\usepackage{color,colortbl}
\usepackage{xcolor}

\definecolor{LightGreen}{rgb}{0.88,1,0.88}
\definecolor{LightGray}{rgb}{0.65,0.65,0.65}

\usepackage{wrapfig}
\usepackage{bm}
\usepackage{amsmath,amsfonts,amsthm}
\usepackage{xspace}
\usepackage{enumitem}
\usepackage[normalem]{ulem}

\usepackage{thmtools} 
\usepackage{thm-restate}

\newcommand{\name}{TRIPODD\xspace}
\newcommand{\firstcl}{black} 
\newcommand{\secondcl}{black} 

\newcommand{\srinivas}[1]{{\color{black} #1}}

\setlist{nolistsep}

\AtBeginDocument{%
  \providecommand\BibTeX{{%
    \normalfont B\kern-0.5em{\scshape i\kern-0.25em b}\kern-0.8em\TeX}}}

\setcopyright{acmcopyright}
\copyrightyear{2018}
\acmYear{2018}
\acmDOI{3632410.3632434}

\acmConference[CODS-COMAD]{Joint International Conference on Data Science and Management of Data}{Jan 04--07,
  2024}{Bengaluru, KA, IND}
  
\acmPrice{15.00}
\acmISBN{978-1-4503-XXXX-X/18/06}

\begin{document}

\title{Interpretable Model Drift Detection}


\author{Pranoy Panda}
\orcid{0009-0006-4457-9909}
\affiliation{%
  \institution{Indian Institute of Technology}
  \city{Hyderabad}
  \country{India}
}

\author{Kancheti Sai Srinivas}
\orcid{0000-0001-9300-7209}
\affiliation{%
  \institution{Indian Institute of Technology}
  \city{Hyderabad}
  \country{India}
}

\author{Vineeth N Balasubramanian}
\orcid{0000-0003-2656-0375}
\affiliation{%
  \institution{Indian Institute of Technology}
  \city{Hyderabad}
  \country{India}
}

\author{Gaurav Sinha}
\orcid{0000-0002-3590-9543}
\affiliation{%
  \institution{Microsoft Research}
  \city{Bengaluru}
  \country{India}
}

\renewcommand{\shortauthors}{Panda, et al.}

\begin{abstract}
  Data in the real world often has an evolving distribution. Thus, machine learning models trained on such data get outdated over time. This phenomenon is called model drift. Knowledge of this drift serves two purposes: (i) Retain an accurate model and (ii) Discovery of knowledge or insights about change in the relationship between input features and output variable w.r.t. the model. Most existing works focus only on detecting model drift but offer no interpretability. In this work, we take a principled approach to study the problem of interpretable model drift detection from a risk perspective using a feature-interaction aware hypothesis testing framework, which enjoys guarantees on test power. The proposed framework is generic, i.e., it can be adapted to both classification and regression tasks. Experiments on several standard drift detection datasets show that our method is superior to existing interpretable methods (especially on real-world datasets) and on par with state-of-the-art black-box drift detection methods. We also quantitatively and qualitatively study the interpretability aspect including a case study on USENET2 dataset. We find our method focuses on model and drift sensitive features compared to baseline interpretable drift detectors.
\end{abstract}

\maketitle

\section{Introduction}
A standard assumption in traditional supervised learning settings is that input data is sampled from a stationary distribution. In reality, however, many application domains -- including finance, healthcare, energy informatics, and communications -- generate data that evolve with time, i.e., they are non-stationary \cite{moons2012risk,davis2017calibration,minne2012effect}. Such an evolution of data over time can make models learned through standard supervised learning techniques underperform, i.e., the decision boundary learned by the model drifts away from the actual decision boundary over time \cite{khamassi2018discussion}. This phenomenon of model degradation is termed as \textit{model drift}. Related settings have been studied under different names in literature -- most primarily, as concept drift -- which we discuss along with related work in Sec \ref{rel_works}. To better understand model drift, consider a machine learning (ML) or neural network (NN) model trained on data points produced by a data-generating process. This generating process can change over time, due to evolving dynamics of a given application setting, resulting in a shift in the covariate distribution (a.k.a \textit{covariate shift}), shift in feature-conditioned output/posterior distribution (a.k.a \textit{posterior shift/real concept drift}), or both \cite{gama2014survey}. 
However, not all data shifts result in significant model performance degradation - i.e. some data shifts are benign w.r.t. the model (for e.g., data that lie far away from the decision boundary in the correct direction). Retraining the model in such scenarios could be unnecessary and rather increase the cost of model deployment. 

Besides retraining when the model performance degrades, it is useful to understand such a drift in terms of input features. It can be helpful to deduce change in variables of importance, i.e. features that were predictive before the drift but not after, and vice versa. With the growing emphasis on interpretable models, such kinds of insights are helpful for users to understand which variable -- say, user demand, geographic location or season -- changed in such a drift. 
Thus, \textit{interpretable model drift detection} is an important problem to be studied as it has a potentially wide range of applications including predictive maintenance \cite{zenisek2019machine}, social media analysis \cite{costa2014concept,muller2020addressing} and malware detection \cite{jordaney2017transcend}, where model drifts are common.


\begin{table*}
\small
  \centering
  \caption{\footnotesize Comparison of our method with different model drift detectors in literature. Marginal \ \cite{dos2016fast}  and Conditional \ \cite{kulinski2020feature}  are covariate shift methods, which do not focus on model drift. 
  $\checkmark$ and $\times$ represent \textit{True} and \textit{False}, respectively.}
  \vspace{-5pt}
    \label{table:comp_interp_methods}
  \begin{tabular}{lccccc}
    \toprule
    Properties $\rightarrow $   & Focus on & Feature-level & Relevant for & Relevant for& No assumptions  \\
    Methods $\downarrow $ & model drift & interpretability & classification task & regression task & on covariates \\
    \midrule
    Marginal \ \footnotesize{\cite{dos2016fast} }  & $\times$  & $\checkmark$ & $\checkmark$ & $\checkmark$  & $\checkmark$  \\
    Conditional \ \footnotesize{\cite{kulinski2020feature} } & $\times$ & $\checkmark$ & $\checkmark$ & $\checkmark$ & $\times$  \\
    \midrule
    MDDM  \footnotesize{\cite{pesaranghader2018mcdiarmid} } & $\checkmark$   & $\times$ & $\checkmark$ & $\times$  & $\checkmark$  \\
    DDM \ \footnotesize{\cite{gama2004learning} } & $\checkmark$    & $\times$ & $\checkmark$ & $\times$  & $\checkmark$  \\
    ADWIN \footnotesize{\cite{bifet2007learning} } & $\checkmark$    & $\times$ & $\checkmark$ & $\checkmark$  & $\checkmark$  \\
    KSWIN \footnotesize{\cite{raab2020reactive} } & $\checkmark$   & $\times$ & $\checkmark$ & $\checkmark$  & $\checkmark$  \\
    \midrule
    \rowcolor{LightGreen} \name (Ours)  & $\checkmark$   & $\checkmark$  & $\checkmark$ & $\checkmark$ & $\checkmark$ \\
    \bottomrule
    \end{tabular}
    \vspace{-5pt}
\end{table*}
Existing methods in literature to detect model drift include KS test based adaptive WINdowing (KSWIN) \cite{raab2020reactive}, McDiarmid Drift Detection Method (MDDM) \cite{pesaranghader2018mcdiarmid}, Drift Detection Method (DDM) \cite{gama2004learning}, Early Drift Detection Method (EDDM) \cite{baena2006early} and Adaptive Windowing Algorithm (ADWIN) \cite{bifet2007learning}. Although these methods detect model drift, they are not interpretable. 
The limited efforts that have considered interpretability in drift detection \cite{kulinski2020feature,dos2016fast} are restricted to drifts in the covariate space, which may not always lead to model drift. To the best of our knowledge, no holistic framework exists to address interpretable model drift detection. Existing post-hoc explainability methods are not intended for distribution shifts, and thus give unreliable explanations under drifts \cite{lakkaraju2020robust}. Therefore, in this work, we propose feature-inTeraction awaRe InterPretable mOdel Drift Detection (\name), a method that leverages hypothesis testing and model risk to detect model drift and simultaneously interpret it w.r.t. input features. Table \ref{table:comp_interp_methods} summarizes the comparison of \name with existing methods, and shows its usefulness and generalizability over earlier efforts.

The proposed \name method adopts a first-principles approach and uses the base model's empirical risk to directly study change in model performance on the prediction task w.r.t. a model class. Empirical risk minimization \cite{vapnik1991principles} has remained a key foundation of the field of machine learning, which is used to train ML and NN models; hence, defining change in decision boundary in terms of model risk is a natural choice. To attain feature-level interpretability of the drift, we formally define feature-sensitive model drift definition (Defn \ref{defn:fs-model-drift}) and construct a hypothesis test around that definition, which is sensitive to feature interactions learned by the underlying NN model. This is then used for detecting and interpreting drifts on real-world datasets. Our key contributions are summarized below: 
\begin{itemize}[leftmargin=*]
    \item We propose a new method, \name, for interpretable model drift detection. To the best of our knowledge, this is the first such work towards feature-interpretable model drift detection, paralleling the significant uptake of interpretability across application domains at this time. 
    
    \item As \name uses only model risk to achieve this objective, it can be applied to both classification and regression tasks, thus making it a fairly generic method.
    
    \item We theoretically analyze the hypothesis testing framework underlying our method, and show that our proposed framework has guarantees on test power.
    
    \item We perform a comprehensive suite of experiments on 10 synthetic and 5 real-world datasets which show the superior interpretability of \name over well-known state-of-the-art baseline methods for the task, while performing at par or better in terms of model drift detection performance. We study interpretability both quantitatively and qualitatively to validate the usefulness of interpretations in our framework. 
\end{itemize}

\vspace{-4pt}
\section{Related Work}
\label{rel_works}
Distribution shift over time and its detection has been studied extensively in literature under different names \cite{tsymbal2004problem,gama2014survey,lu2018learning, sato2021survey, gonccalves2014comparative, patil2021concept,agrahari2021concept}. 
The most popular among them is \textit{concept drift} \cite{gama2014survey,tsymbal2004problem}. Such a drift could occur due to change in covariate distribution (referred to as \textit{virtual drift} or \textit{covariate shift}), or change in the posterior distribution (referred to as \textit{real concept drift} or \textit{posterior shift}) \cite{gama2014survey}, or both. \textit{Dataset shift} \cite{moreno2012unifying} is another term used to capture such shift in distribution. 
In this work, we study model drift, which deals with deterioration of the model performance due to evolution in the data distribution. It does not directly fall into the above mentioned categories as data distribution shift (covariate shift or posterior shift) need not always lead to significant model degradation. \\
Below we give a summary of works in the space of concept drift detection methods and research efforts for interpreting drifts. 

\noindent \textbf{Concept Drift Detection Methods:} Methods such as LSDD-CDT \cite{bu2016pdf}, Marginal \cite{dos2016fast} and Conditional Test \cite{kulinski2020feature} study covariate shift. Focusing on the covariate distribution can make these methods vulnerable to benign drifts in real-world data streams that do not change the model. These methods are hence known to generate many false positives when applied, thus limiting their usefulness in practice. On the other hand, posterior shift methods (also called real concept drift methods) such as MDDM \cite{pesaranghader2018mcdiarmid}, DDM \cite{gama2004learning}, EDDM \cite{baena2006early} and ADWIN \cite{bifet2007learning} track misclassifications of a classifier and detect drift when the distribution of error changes, i.e. track change in posterior distribution by using cues from a given model. Such methods are relatively more robust to benign drifts in data streams. 
We study model drift instead, which is more contemporarily practical but does not fall into these categories as we take the model perspective instead of the data perspective.

\noindent \textbf{Interpretability in Drift Detection:} From an interpretability perspective, while most popular drift detection methods including MDDM \cite{pesaranghader2018mcdiarmid}, DDM \cite{gama2004learning}, EDDM \cite{baena2006early}, \srinivas{SDDM \cite{micevska2021sddm}}, ADWIN \cite{bifet2007learning} and KSWIN \cite{raab2020reactive} are black-box methods, there have been efforts that attempt to detect drift and simultaneously understand different aspects of the drift that can provide a user with insights. These can be broadly categorized into visualization-based methods and feature-interpretable methods. \textit{Visualization-based methods} provide insights by maps or plots which inform the user about the distribution change. For e.g., \cite{webb2018analyzing} studied drift using quantitative descriptions of drift in the marginal distributions, and then used marginal drift magnitudes between time periods to plot heat maps that gives insights of the drift. \cite{pratt2003visualizing} developed a visualization tool that used parallel histograms to study concept drift that a user could visually inspect. 

\textit{Feature-interpretable methods} \cite{dos2016fast,kulinski2020feature,demvsar2018detecting,lobo2021curie} attempt to detect drift and simultaneously notify which features might be responsible for it. 
Marginal \cite{kifer2004detecting, dos2016fast} performs a feature-wise Kolmogorov-Smirnov (KS) test, while Conditional Test \cite{kulinski2020feature} performs a hypothesis test to check for a change in the distribution of each feature conditioned on the other input features. \cite{kulinski2020feature} addresses the adversarial drift detection problem in their work which is slightly different from the standard drift detection problem. \cite{lobo2021curie} introduced a cellular automaton-based drift detector, where its cellular structure becomes a representation of the input feature space. However, this work does not deal with model drift. 
While our method also falls in this category of feature-interpretable methods, our objective of model drift differs from these methods. 
Using model risk as a direct indicator allows us to apply our method in all kinds of prediction tasks, including classification and regression. A relatively older method \cite{harel2014concept} also uses a notion of model risk to study concept drift, but does not address interpretability. Our feature-sensitive model drift definition (Sec \ref{methodology}) allows us to model the drift in an interpretable manner \& provide theoretical guarantees that are desirable for a hypothesis testing framework.

\section{Background and Preliminaries}
\label{sec_bgd}
\paragraph{Notations.} We use capital letters to represent random variables and corresponding small letters to represent the values taken by these random variables. Bold-faced letters denote vectors or sets of variables. For each positive integer $n$, we denote the set $\{1,\ldots,n\}$ by $[n]$ and the set $\{m,m+1,...,n\}$ by $[m,n]$. By $x\sim p(X)$, we mean that $x$ is obtained by sampling from the distribution $p(X)$.
We represent our input covariates by $\bm{X}\in \mathcal{X}$ where $\mathcal{X} \subset \mathbb{R}^d$, and output variable as $Y$, which takes values in set $\mathcal{Y}$, such that $\mathcal{Y}\subset \mathbb{R}$ for regression problems and $\mathcal{Y}=[C]$ for classification problems ($C$ is the number of classes).
Let $D_p = \{x_i,y_i\}_{i=1}^{T} \sim p(X,Y)$ and $D_q = \{x_i,y_i\}_{i=T+1}^{N} \sim q(X,Y)$ be set of two samples and $\mathcal{H}$ be some model class. The risk associated with a model $h \in \mathcal{H}$ on distribution $p(\bm{X},Y)$ is given by $\mathcal{R}_{p}^L(h) = \mathbb{E}_{(\bm{x},y) \sim p(\bm{X},Y) }[L(h(x),y)]$, for some loss function $L$. The corresponding empirical risk for sample $D_p$ is given by $\hat{\mathcal{R}}_{D_p}^L(h) = \frac{1}{T}\sum\limits_{i=1}^{T}L(h(x_i),y_i)$. Unless otherwise specified, we will assume the loss function to be some fixed unknown function and use $\mathcal{R}_p(h), \hat{\mathcal{R}}_{D_p}(h)$ instead of $\mathcal{R}_{p}^L(h), \hat{\mathcal{R}}_{D_p}^L(h)$ respectively. For each $i\in [d]$, we define $\bm{e}_i$ to be the vector with $1$ in the $i^{th}$ co-ordinate and $0$ elsewhere. For any set $S\subset [d]$ and vector $\bm{x}\in \mathbb{R}^d$, by $\bm{x}\odot S$ we denote the orthogonal projection of $\bm{x}$ onto the subspace spanned by $\{\bm{e}_i, i\in S\}$ i.e. $\bm{x}\odot S$ matches $\bm{x}$ at all co-ordinates belonging to $S$ and is $0$ in all the other co-ordinates. 

For any subset $S\subset [d]$, we define the subset-specific risk as $\mathcal{R}^{S}_{p}(h) = \mathbb{E}_{(\bm{x},y) \sim p }[L(h(x\odot S),y)]$. Finally, for each $i\in [d]$, we define a vector $\Delta_p^i(h) \in \mathbb{R}^{2^d}$ that captures the change in risk when the $i^{th}$ feature is added to all the other subsets of features. Using the index of all subsets $S\subset[d]$ (in any fixed order), we formally define the co-ordinates of the $2^d$ dimensional vector $\Delta _{p}^{i} (h)$ as:
\begin{equation*}
    (\Delta _{p}^{i} (h))_S = \mathcal{R}^{S}_{p}(h) -\mathcal{R}^{S\cup \{i\}}_{p}(h)    
\end{equation*}
Using these notations, we now define model drift and feature-sensitive model drift to study the problem of interpretable model drift detection in a principled manner. We note that our definitions can be used for both classification and regression settings since they depend only on the change in the risk
of the model.  


\begin{definition}[Model Drift]
\label{defn:model-drift}
Consider a stream of samples $\{(\bm{x}_t, y_t): t=1,2,\ldots\}$. We say a model drift occurs at time $t=T$, if there exist two distributions $p(\bm{X},Y), q(\bm{X},Y)$ on the joint variable $(\bm{X}, Y)$ and a model $h$ trained on samples from $p(\bm{X},Y)$ distribution, s.t.,
\begin{enumerate}
    \item $(\bm{x}_t,y_t)\sim p(\bm{X},Y)$ for $t\leq T$,
    \item $(\bm{x}_t,y_t)\sim q(\bm{X},Y)$ for $t> T$, and
    \item $\mathcal{R}_{p}(h)\neq \mathcal{R}_{q}(h)$.
\end{enumerate}
\end{definition}

The above definition says that there is a model drift if the true risk of the model $h$ is different when the distribution changes from $p(\bm{X},Y)$ to $q(\bm{X},Y)$. 
Change in risk indicates that the model may not perform as well on samples from the new distribution $q(\bm{X},Y)$. 

\begin{definition}[Feature-Sensitive Model Drift]
\label{defn:fs-model-drift}
Consider a stream of samples $\{(\bm{x}_t, y_t): t=1,2,\ldots\}$. We say that a feature-sensitive model drift occurs at time $t=T$, if there exist distributions $p(\bm{X},Y),$ $q(\bm{X},Y)$
on the joint variable $(\bm{X}, Y)$ and a model $h$ trained on samples from $p(\bm{X},Y)$, such that:
\begin{enumerate}
    \item $(\bm{x}_t,y_t)\sim p(\bm{X},Y)$ for $t\leq T$,
    \item $(\bm{x}_t,y_t)\sim q(\bm{X},Y)$ for $t> T$, and
    \item There exists $i \in [d]$, such that $\Delta _{p}^{i}(h) \neq \Delta _{q}^{i}(h)$. 
    
    
    
    
\end{enumerate}
\end{definition}

As mentioned earlier, $\Delta_p^i(h)$ contains the change in subset specific risk $\mathcal{R}_p^S(h)$ for all subsets $S\subset [d]$, when the $i^{th}$ feature is added to it. Intuitively, it contains the ``impact'' of adding the $i^{th}$ feature to other subsets of features, measured as a change in the subset specific risk. Thus, a feature sensitive drift occurs if, for some feature, this impact is different for $t\leq T$ and $t>T$. Such an approach allows us to consider all possible feature interactions, and is hence reliable. One could view our approach as a first-principles approach premised on model risk and feature-level interpretability. 
We now describe our hypothesis testing framework for detecting feature-sensitive model drift. 

\section{\name: Methodology}
\label{methodology}
We begin by defining the hypothesis testing framework which we build on Definition \ref{defn:fs-model-drift}. We then define our test statistic which helps to conduct our hypothesis test on samples, followed by our overall methodology of \name including its algorithm. We also theoretically analyze the test power of our test statistic and show that it converges to $1$ as the number of samples $n\rightarrow \infty$.

\vspace{-8pt}
\subsection{Hypothesis Testing Framework}
\label{subsec:hyp}
As stated earlier, our framework is directly built on top of the definition of feature-sensitive model drift definition (Defn \ref{defn:fs-model-drift}), as it relates existence of model drift to input features. It states that the presence of such a drift requires at least one feature $k$ and a subset of features $S$ for which the change in subset-specific risk $\mathcal{R}_p^S(h) - \mathcal{R}_p^{S\cup\{k\}}(h)$ and  $\mathcal{R}_q^S(h) - \mathcal{R}_q^{S\cup\{k\}}(h)$ are different. These features ($k$) have different effects on model risk for the two distributions $p$ and $q$ in Defn \ref{defn:fs-model-drift}, and can thus be used as an interpretation of the drift. Our method leverages a hypothesis testing framework to identify these features.
\begin{definition}[Hypothesis Test]
\label{defn:hypothesis2}
The null hypothesis \textbf{H}$_0$ and alternate hypotheses \textbf{H}$_a$ for the effect of the $k^{th}$ feature on model risk is given by: 
\paragraph*{\textbf{H}$_0$:} 
For all subsets of features $S\subseteq [d]$, 
\begin{equation*}
    \mathcal{R}^{S}_{p}(h) - \mathcal{R}^{S\cup \{k\}}_{p}(h) = \mathcal{R}^{S}_{q}(h) - \mathcal{R}^{S\cup \{k\}}_{q}(h)
\end{equation*}
\paragraph{\textbf{H}$_a$:} 
$\exists S \subseteq [d]$, such that,
\begin{equation*}
    \mathcal{R}^{S}_{p}(h) - \mathcal{R}^{S\cup \{k\}}_{p}(h) \neq \mathcal{R}^{S}_{q}(h) - \mathcal{R}^{S\cup \{k\}}_{q}(h)
\end{equation*}
\end{definition}
We signal a drift if and only if \textbf{H}$_0$ is rejected for some feature $k\in [d]$. We note that the null hypothesis \textbf{H}$_0$ is true if and only if there is no feature-sensitive model drift, and, \textbf{H}$_a$ is true otherwise.

Let $d^k(h) \coloneqq\max_{S\subseteq F} | (\mathcal{R}^{S}_{p}(h) - \mathcal{R}^{S\cup \{k\}}_{p}(h)) - (\mathcal{R}^{S}_{q}(h) - \mathcal{R}^{S\cup \{k\}}_{q}(h)) |$ and $\widehat{d}^k(h)$ be its sample estimate i.e., $\widehat{d}^k(h) \coloneqq \max_{S\subseteq F} | (\hat{\mathcal{R}}^{S}_{D_p}(h) - \hat{\mathcal{R}}^{S\cup \{k\}}_{D_p}(h)) - (\hat{\mathcal{R}}^{S}_{D_q}(h) - \hat{\mathcal{R}}^{S\cup \{k\}}_{D_q}(h))|$. It is easy to see that $d^k(h)>0$ (strictly greater) if and only if the alternate hypothesis is true, otherwise it is $0$. We define our test statistic as follows.

\begin{definition}[Test Statistic]
\label{def:statistic}
Given two streams of samples $D_p = \{(\bm{x}_i, y_i)\sim p(\textbf{X},Y): i\in [n]\}$ and $D_q = \{(\bm{x}_j, y_j)\sim q(\textbf{X},Y):$ $j\in [n+1,2n]\}$, along with a model $h$ trained on samples from $p(\bm{X}, Y)$, we define our sample test statistic $\widehat{c}^k_n(h)$ for the $k^{th}$ feature as:
\begin{equation}
\label{eqn:statistic}
    \widehat{c}^k_n(h)\coloneqq n\widehat{d}^k(h) 
\end{equation}
The population counterpart is defined as $c_{n}^k(h) \coloneqq nd^k(h)$.
\end{definition}

One could view the quantity $d^k(h)$ used in our test statistic above as similar to the Marginal Contribution Importance (MCI) score in \cite{catav2021marginal} (or Shapley values \cite{shapely1953value} in terms of measuring feature contributions); however, MCI or Shapley values do not focus on model drift. Besides, our statistic and our corresponding hypothesis testing framework directly follow from our definitions in Defns \ref{defn:fs-model-drift} and \ref{defn:hypothesis2} towards model drift detection, making this an equivalent first-principles approach for interpreting model drift detection. 

\subsection{Methodology}
\label{desc_algo}
Given the test statistic defined in Defn \ref{eqn:statistic}, we now describe our methodology for drift detection. \name uses a sliding window procedure and maintains a reference window (samples from old distribution), new samples window (samples from current distribution) and a model trained on samples from old distribution. It compares the risk between reference ($Z_R$) and new samples windows ($Z_N$) for every feature by using the test statistic stated above (note that first $n - \lfloor nr \rfloor$ samples of $Z_N$ are used for the test to ensure risk is computed on $Z_R$ \& $Z_N$ on an equal number of samples). In any hypothesis testing framework, it becomes important to understand the statistical significance of the test statistic values at a given point. To this end, we use thresholds evaluated using a bootstrapping procedure, briefly described below. If there exists at least one feature for which the computed test statistic is greater than the corresponding threshold, then it declares model drift and all features that result in a drift become part of the drift interpretation. The entire procedure for implementing \name is detailed in Algorithm \ref{alg:sup}. The $Performance$ variable stores the model performance (accuracy in classification, or $R^2$ value in regression) across the data stream.

\textit{Bootstrap procedure:} (for $K$ bootstraps \& significance level $\alpha$) In this procedure we merge and shuffle samples from $Z_N$ and $Z_R$, and then pick $K$ two-samples from this mixture. This simulates the null hypothesis. Finally, we calculate the test statistic for these $K$ two-samples and return the (1- $\frac{\alpha}{d}$)-th quantile of the test statistic values as our threshold (we use Bonferroni correction as we perform multiple comparisons). 

\setlength{\textfloatsep}{2pt}
\begin{algorithm}[t]
\footnotesize
\caption{\footnotesize \name (feature-inTeraction  awaRe  InterPretable mOdel Drift Detection)}\label{alg:sup}
\begin{algorithmic}[1]
\Require $\mathcal{H}, n, \alpha, r, K, \delta$, $\mathcal{Z} = \{\bm{z}_t = (\bm{x}_t, y_t) : t=1,2,\ldots\}$ 
\State Create empty list \textit{Performance} $\gets \phi$.
\Comment{Model performance across stream}
\State Create empty list \textit{Interpretation} $\gets \phi$. \Comment{Drift related features}
\State $\tilde{n} \gets n - \lfloor nr \rfloor$ \Comment{Effective window size}
\State Initialize $i \gets n$, and model $h \gets$ \Call{GetModel}{$\mathcal{H}, \{\bm{z}_t : t\in \{1,\ldots, \lfloor nr \rfloor\}$} 
\State $\mathcal{Z}_R = \{\bm{z}_t : t\in \{1+\lfloor nr \rfloor,\ldots, n\}\}$.

\While{$ True $}
\State Flag $ \gets False$ \Comment{Drift flag}
\State $\mathcal{Z}_N = \{\bm{z}_t : t\in [i+1,\dots, i+n]\}$.
\State Compute $\hat{c}_n^k(h)$, $\forall k\in [d]$, using $\mathcal{Z}_R, \mathcal{Z}_N, h$ in Eq \ref{eqn:statistic}.
\State Get thresholds $(T_{\alpha}^1,\ldots,T_{\alpha}^d) \gets $ \Call{Bootstrap}{$h,\alpha,K,\mathcal{Z}_R,\mathcal{Z}_N$}
    \If{for any $k\in [d]$, $\hat{c}_n^k(h) > T_{\alpha}^k$}
    \State Flag $\gets True$, Add all such $k$s to list \textit{Interpretation}.
    \EndIf
\If{Flag $ = True$} \Comment{ Drift detected}
    \State $h \gets \Call{GetModel}{\mathcal{H}, \{\bm{z}_t : t\in [i+1,\dots, i+ \lfloor nr \rfloor]}$ 
    \State $\mathcal{Z}_R \gets \{\bm{z}_t : t\in [i+ \lfloor nr \rfloor+1,\dots,i+n]\}$ 
    \State Add model perf of $h$ on $Z_R$ to the \textit{Performance} list
    \State $i  \gets  i + n $ \Comment{ Shift windows by $n$}
    \State Print the list \textit{Interpretation} and reset it to $\phi$.
\Else
    \State Add model perf of $h$ on $Z_N$ to \textit{Performance} list
    \State $i  \gets i + \delta$ \Comment{Shift windows by $\delta$ if no drift detected}
\EndIf
\EndWhile
\end{algorithmic}
\end{algorithm}

\textbf{Efficient Treatment of Subsets.} For dealing with datasets with a large number of features, following earlier efforts such as MCI \cite{catav2021marginal} that leverage sampling techniques, we use the random sampling technique in \cite{covert2020understanding} 
to reduce the computational overhead. A random set of permutations of features is sampled, for which our test statistic is computed. We show in our experiments and analysis (see Section \ref{ablation} that our method's time complexity is significantly lesser than the sampling rates of the datasets typically used for model drift detection, making this application setting a relevant one for such a feature-interaction based approach.
\vspace{-4pt}
\subsection{Guarantees of Test Statistic}
To show the goodness of the proposed test statistic, we theoretically analyze its test power and consistency. We use a bootstrap sampling approach from \cite{efron1992bootstrap} to simulate the null hypothesis which addresses the consistency of our test. We now show that the power of our test converges i.e. for any threshold (corresponding to some significance level $\alpha$), the probability that our statistic is larger than threshold tends to 1, when the alternate hypothesis is true. We formally state our results below and provided the proof in Appendix. 

\vspace{-5pt}
\begin{restatable}[Convergence of Test Power]{theorem}{consistencyandtestpower}
\label{thm:cons_and_test_power}
The test power of the proposed hypothesis test in Definition \ref{defn:hypothesis2} converges to the ideal value 1 under the alternate hypothesis, i.e. suppose the alternate hypothesis $H_a$ is true for some feature $k\in [d]$, then, for any $t>0$,
$\lim _{n\rightarrow \infty } \mathbb{P}[\widehat{c}^{k}_{n}(h) > t] = 1$.
\end{restatable}


\vspace{-10pt}
\section{Experiments and Results}

\begin{table*}[t]
  \setlength{\tabcolsep}{3pt} 
  \centering
  \caption{\small \textit{Drift Detection Results:} We report Average Model Performance across the data stream ($M$) for all datasets, and Precision ($P_{det}$) + Recall ($R_{det}$) for synthetic datasets with ground truth drift locations. ($M$ = accuracy for classification; $M$ = $R^2$ value for regression). We compare \name with 2 interpretable (Marginal \& Conditional) and 4 black-box drift detectors (KSWIN, MDDM, DDM, ADWIN) on 10 synthetic datasets and 5 real-world datasets. Interpretable methods are \fcolorbox{gray!22}{gray!22}{highlighted in gray}. \textit{AD (Always Drift)} for Marginal \& Conditional indicates that these methods, since not designed for model drift, detected drift at every new window, making them ineffective for drift detection. Dataset variations with \textit{imbalance} are added to show results with increased complexity of class imbalance. Higher is better for all metrics. Values in \textbf{bold} \& \underline{underlined} indicate best and second best.}
  \label{table:det_results}
  \scalebox{0.95}{
  \begin{tabular}{l|>{\columncolor[gray]{0.93}}c|>{\columncolor[gray]{0.93}}c|>{\columncolor[gray]{0.93}}c|c|c|c|c}
    \toprule
    \textbf{Methods} $\rightarrow$ & \name (Ours) & Marginal & Conditional & KSWIN & MDDM & DDM & ADWIN\\
    \textbf{Datasets} $\downarrow$ & \small{M \ \textcolor{\secondcl}{$P_{det}$ \ $R_{det}$}} & \small{M \ \textcolor{\secondcl}{$P_{det}$ \ $R_{det}$}} & \small{M \ \textcolor{\secondcl}{$P_{det}$ \ $R_{det}$}} & \small{M \ \textcolor{\secondcl}{$P_{det}$ \ $R_{det}$}} & \small{M \ \textcolor{\secondcl}{$P_{det}$ \ $R_{det}$}} & \small{M \ \textcolor{\secondcl}{$P_{det}$ \ $R_{det}$}} & \small{M \ \textcolor{\secondcl}{$P_{det}$ \ $R_{det}$}} \\
    \midrule
    \multicolumn{8}{l}{\textit{Classification Task}} \\
    \midrule 
    Sine (balance) & \textbf{55.5} \ \ \textcolor{\secondcl}{\textbf{1.0} \ \ \textbf{1.0}} &
      42.2 \ \ \textcolor{\secondcl}{0.0 \ \ 0.0}  & 
      42.2 \ \ \textcolor{\secondcl}{0.0 \ \ 0.0}  &
      51.3  \ \ \textcolor{\secondcl}{\textbf{1.0} \ \ \textbf{1.0}} &
      \underline{55.2} \ \ \textcolor{\secondcl}{\textbf{1.0} \ \ \textbf{1.0}} &
      50.0  \ \ \textcolor{\secondcl}{\textbf{1.0} \ \ \textbf{1.0}} &
      \textbf{55.5} \ \ \textcolor{\secondcl}{\textbf{1.0} \ \ \textbf{1.0}}  \\
    
    Sine (imbalance) & 55.0 \ \ \textcolor{\secondcl}{\textbf{1.0} \ \ \textbf{1.0}} &
      42.2 \ \ \textcolor{\secondcl}{0.0 \ \ 0.0} & 
      42.2 \ \ \textcolor{\secondcl}{0.0 \ \ 0.0} &
      \textbf{59.4} \ \ \textcolor{\secondcl}{\textbf{1.0} \ \ \textbf{1.0}} &
      \underline{59.0} \ \ \textcolor{\secondcl}{\textbf{1.0} \ \ \textbf{1.0}} &
      50.0 \ \ \textcolor{\secondcl}{\textbf{1.0} \ \ 0.3} &
      56.0 \ \ \textcolor{\secondcl}{\textbf{1.0} \ \ \textbf{1.0}}\\
    
    Agrawal (balance) & \textbf{55.0} \ \ \textcolor{\secondcl}{\textbf{1.0} \ \ \textbf{1.0}} &
      \textbf{55.0} \ \ \textcolor{\secondcl}{0.8 \ \ \textbf{1.0}}  &
      \underline{54.0} \ \ \textcolor{\secondcl}{\textbf{1.0} \ \ \textbf{1.0}} &
      50.3 \ \ \textcolor{\secondcl}{\textbf{1.0} \ \ \textbf{1.0}} &
      51.5  \ \ \textcolor{\secondcl}{\textbf{1.0} \ \ \textbf{1.0}} &
      50.0  \ \ \textcolor{\secondcl}{\textbf{1.0} \ \ \textbf{1.0}} &
      51.1  \ \ \textcolor{\secondcl}{\textbf{1.0} \ \ \textbf{1.0}}\\
    
    Agrawal (imbalance) & \underline{52.0} \ \ \textcolor{\secondcl}{\textbf{1.0} \ \ \textbf{1.0}} &
      \underline{52.0} \ \ \textcolor{\secondcl}{0.8 \ \ 0.8}  &
      \underline{52.0}  \ \ \textcolor{\secondcl}{\textbf{1.0} \ \ 0.5} &
      48.0 \ \ \textcolor{\secondcl}{0.8 \ \ \textbf{1.0}} &
      \underline{52.0}  \ \ \textcolor{\secondcl}{\textbf{1.0} \ \ 0.8} &
      45.0 \ \ \textcolor{\secondcl}{\textbf{1.0} \ \ 0.3} &
      \textbf{56.2}  \ \ \textcolor{\secondcl}{\textbf{1.0} \ \ \textbf{1.0}}\\
    
    Mixed & \textbf{99.2} \ \ \textcolor{\secondcl}{0.7 \ \ \textbf{1.0}} &
      98.9 \ \ \textcolor{\secondcl}{0.5 \ \ \textbf{1.0}}  &
      \textbf{99.2} \ \ \textcolor{\secondcl}{0.5 \ \ \textbf{1.0}}  &
      98.1 \ \ \textcolor{\secondcl}{\textbf{1.0} \ \ 0.3} &
      98.1 \ \ \textcolor{\secondcl}{\textbf{1.0} \ \ 0.3} &
      98.0 \ \ \textcolor{\secondcl}{\textbf{1.0} \ \ 0.4} &
      \underline{99.0} \ \ \textcolor{\secondcl}{0.8 \ \ \textbf{1.0}}\\    
    
    Aug-Mixed   & \textbf{94.0} \ \ \textcolor{\secondcl}{\textbf{1.0} \ \ \textbf{0.8}} &
      90.0 \ \ \textcolor{\secondcl}{0.5 \ \ 0.6}  &
      90.0 \ \ \textcolor{\secondcl}{0.4 \ \ 0.6}  &
      \underline{92.7} \ \ \textcolor{\secondcl}{\textbf{1.0} \ \ 0.6} &
      \underline{92.7} \ \ \textcolor{\secondcl}{\textbf{1.0} \ \ 0.6} &
      90.0 \ \ \textcolor{\secondcl}{\textbf{1.0} \ \ 0.4} &
      92.0 \ \ \textcolor{\secondcl}{\textbf{1.0} \ \ \textbf{0.8}}\\

    SEA  & \textbf{92.0} \ \ \textcolor{\secondcl}{\underline{0.7} \ \ \textbf{1.0}} &
      88.6 \ \ \textcolor{\secondcl}{\textbf{1.0} \ \ 0.3}  &
      88.2 \ \ \textcolor{\secondcl}{\textbf{1.0} \ \ 0.3}  &
      \underline{90.0} \ \ \textcolor{\secondcl}{\textbf{1.0} \ \ 0.7} &
      \underline{90.0} \ \ \textcolor{\secondcl}{\textbf{1.0} \ \ 0.7} &
      89.0 \ \ \textcolor{\secondcl}{\textbf{1.0} \ \ 0.3} &
      89.5 \ \ \textcolor{\secondcl}{\textbf{1.0} \ \ 0.3}\\

    SEA-Gradual   & \textbf{88.0} \ \ \ \ - \ \ \ \ \ - \ \ &
      87.3 \ \ \ \ - \ \ \ \ \ - \ \  &
      87.3 \ \ \ \ - \ \ \ \ \ - \ \  &
      \underline{87.8} \ \ \ \ - \ \ \ \ \ - \ \  &
      87.3 \ \ \ \ - \ \ \ \ \ - \ \  &
      87.3 \ \ \ \ - \ \ \ \ \ - \ \  &
      87.3 \ \ \ \ - \ \ \ \ \ - \ \  \\    
    
    Hyperplane   & \textbf{88.1} \ \ \ \ - \ \ \ \ \ - \ \  &
      86.6 \ \ \ \ - \ \ \ \ \ - \ \  &
      87.4 \ \ \ \ - \ \ \ \ \ - \ \  &
      \underline{87.5} \ \ \ \ - \ \ \ \ \ - \ \  &
      86.7 \ \ \ \ - \ \ \ \ \ - \ \  &
      85.9 \ \ \ \ - \ \ \ \ \ - \ \  &
      \underline{87.5} \ \ \ \ - \ \ \ \ \ - \ \ \\

    Airlines   & 57.2 \ \ \ \ - \ \ \ \ \ - \ \  &
      \textcolor{\firstcl}{AD} \ \ \ \ - \ \ \ \ \ - \ \  &
      \textcolor{\firstcl}{AD} \ \ \ \ - \ \ \ \ \ - \ \  &
      \textbf{59.7} \ \ \ \ - \ \ \ \ \ - \ \  &
      56.7 \ \ \ \ - \ \ \ \ \ - \ \  &
      56.5 \ \ \ \ - \ \ \ \ \ - \ \  &
      \underline{57.8} \ \ \ \ - \ \ \ \ \ - \ \ \\

    Electricity   & \textbf{77.6} \ \ \ \ - \ \ \ \ \ - \ \  &
      \textcolor{\firstcl}{AD} \ \ \ \ - \ \ \ \ \ - \ \  &
      \textcolor{\firstcl}{AD} \ \ \ \ - \ \ \ \ \ - \ \  &
      \underline{74.7} \ \ \ \ - \ \ \ \ \ - \ \  &
      74.6 \ \ \ \ - \ \ \ \ \ - \ \  &
      73.4 \ \ \ \ - \ \ \ \ \ - \ \  &
      74.5 \ \ \ \ - \ \ \ \ \ - \ \  \\ 
    
    Weather   & \textbf{77.4} \ \ \ \ - \ \ \ \ \ - \ \  &
      \textcolor{\firstcl}{AD} \ \ \ \ - \ \ \ \ \ - \ \  &
      \textcolor{\firstcl}{AD} \ \ \ \ - \ \ \ \ \ - \ \  &
      75.6 \ \ \ \ - \ \ \ \ \ - \ \  &
      \textbf{77.4} \ \ \ \ - \ \ \ \ \ - \ \  &
      \textbf{77.4} \ \ \ \ - \ \ \ \ \ - \ \  &
      \textbf{72.1} \ \ \ \ - \ \ \ \ \ - \ \ \\
    
    Powersupply   & \textbf{72.6} \ \ \ \ - \ \ \ \ \ - \ \  &
      \textcolor{\firstcl}{AD} \ \ \ \ - \ \ \ \ \ - \ \  &
      \textcolor{\firstcl}{AD} \ \ \ \ - \ \ \ \ \ - \ \  &
      \underline{72.3} \ \ \ \ - \ \ \ \ \ - \ \  &
      \underline{72.3} \ \ \ \ - \ \ \ \ \ - \ \  &
      71.7 \ \ \ \ - \ \ \ \ \ - \ \  &
      72.0 \ \ \ \ - \ \ \ \ \ - \ \  \\
    
    \midrule
    \multicolumn{8}{l}{\textit{Regression Task}} \\
    \midrule 
    Friedmann   & \textbf{0.65} \ \textcolor{\secondcl}{\underline{0.40} \ \ \textbf{1.0}} &
      0.60 \ \textcolor{\secondcl}{\underline{0.4} \ \ \textbf{1.0}} &
      \underline{0.64} \ \textcolor{\secondcl}{0.3 \ \ \textbf{1.0}} &
      0.62 \ \textcolor{\secondcl}{0.3 \ \ \textbf{1.0}} &
      \ \ - \ \ \ \ - \ \ \ \ \ - &
      \ \ - \ \ \ \ - \ \ \ \ \ - &
      0.62 \ \textcolor{\secondcl}{\textbf{0.8} \ \ \textbf{1.0}} \\
    
    Air Quality   & \textbf{0.48} \ \ \ \ - \ \ \ \ \ - \ \  &
      \textcolor{\firstcl}{AD} \ \ \ \ - \ \ \ \ \ - \ \  &
      \textcolor{\firstcl}{AD} \ \ \ \ - \ \ \ \ \ - \ \  &
      \underline{0.22} \ \ \ \ - \ \ \ \ \ - \ \  &
      \ \ - \ \ \ \ - \ \ \ \ \ - &
      \ \ - \ \ \ \ - \ \ \ \ \ - &
      \underline{0.22} \ \ \ \ - \ \ \ \ \ - \ \ \\
    \bottomrule
  \end{tabular}}
  \vspace{-10pt}
\end{table*}

\begin{wraptable}{r}{5cm}
  \vspace{-15pt}
  \setlength{\tabcolsep}{2pt}
\footnotesize
    \centering
    \caption{\footnotesize Datasets used in our experiments. \textit{D} = number of features; \textit{Drift-type} indicates if drift is abrupt or gradual and the type of drift in the dataset -- covariate shift (\textit{cov}), posterior shift (\textit{pos}) or a mixture of both (\textit{mix}).}   
    \vspace{-5pt}
    \label{table:dataset}
    \scalebox{0.95}{
    \begin{tabular}{ccccc}
    \toprule
        \textbf{Type} & \hspace{10pt}\textbf{ Dataset Name} & \textbf{D} & \textbf{Task} & \textbf{Drift-type}\\
        \midrule
        Syn & Sine~\cite{gama2004learning} & 4 & Class. & abrupt,mix\\
        ~ & Agrawal~\cite{agrawal1993database} & 9 & Class. & abrupt,mix\\
        ~ & Mixed~\cite{pesaranghader2016framework} & 6 & Class. & abrupt,cov\\
        ~ & Aug-Mixed & 6 & Class. & abrupt,mix\\
        ~ & SEA~\cite{bifet2010moa} & 3 & Class. & abrupt,mix\\
        ~ & SEA-Gradual~\cite{bifet2010moa} & 3 & Class. & gradual,mix\\
        ~& Hyperplane~\cite{bifet2010moa} & 10 & Class. & gradual,mix \\ 
        ~ & Friedmann~\cite{friedman1991multivariate} & 4 & Reg. & abrupt,mix\\ 
        \midrule
        Real & \ Airlines~\cite{bifet2010moa}  & 7 & Class. & unknown\\
        ~ & 
        USENET2~\cite{harries1999splice} & 100 & Class. & unknown\\
        ~ &
        Electricity~\cite{harries1999splice} & 5 & Class. & unknown\\
        ~ & Weather~\cite{elwell2011incremental} & 9 & Class. & unknown\\
        ~ & PowerSupply~\cite{dau2019ucr}  & 2 & Class. & unknown\\
        ~ & Air Quality~\cite{de2008field} & 8 & Reg. & unknown\\
        \bottomrule \\
    \end{tabular}}
    \vspace{-27pt}
\end{wraptable}
We study two aspects of \name\ -- its model-drift detection capability and its interpretability. We comprehensively evaluate our method on an extensive suite of synthetic, semi-synthetic, and real-world datasets, and observe that \name provides a strong sense of interpretability while performing at par or better than existing state-of-the-art in drift detection capabilities.
\vspace{-7pt}
\subsection{Model Drift Detection}
\label{subsec_drift_detection_results}
We begin with our study of drift detection performance of \name against existing methods.


\noindent \textbf{Datasets:} We evaluate \name on $10$ synthetic datasets and $5$ real-world datasets that are popularly used for drift detection~\cite{lu2018learning}, as listed in Table \ref{table:dataset}. \name is task-agnostic and can detect different model drifts including those caused by covariate shift, posterior shift and a mixture of both. 
We create two harder datasets -- \textit{Sine (imbalance)} and \textit{Agrawal (imbalance)} by adding class imbalance to \textit{Sine} \& \textit{Agrawal} respectively. \srinivas{We expect that data with class imbalance is challenging to drift detectors, as they may ignore drift pertaining to minority classes.} We also added an \textit{Aug-Mixed} dataset that has both covariate and posterior shift by extending the \textit{Mixed} dataset that only has covariate shift. 

\noindent \textbf{Baselines:} 
We compare our method against 6 well-known drift detection methods: KSWIN~\cite{raab2020reactive}, MDDM~\cite{pesaranghader2018mcdiarmid}, DDM~\cite{gama2004learning}, ADWIN~\cite{bifet2007learning}, Marginal~\cite{kifer2004detecting,dos2016fast} and Conditional~\cite{kulinski2020feature}. MDDM and DDM can only detect drifts in the classification setting while KSWIN and ADWIN work in both classification and regression settings. All four methods do not offer feature-level interpretability for the drift. Marginal and Conditional are interpretable by design, but only look at covariate shift (as in Table \ref{table:comp_interp_methods}).

\noindent \textbf{Metrics:} For synthetic datasets where the true drift time is known, following \cite{yu2018request}, we report precision ($P_{det}$) \& recall ($R_{det}$) of detected drifts. True positive (TP), false positive (FP), and false negative (FN) are identified as follows: a detected drift is a TP if it is detected within a small fixed time range (tolerance window) of the true drift location; FN refers to missing a drift within the fixed time range; FP is a detection outside the fixed time range or an extra detection in the fixed time range of the true drift location. The tolerance window for all synthetic datasets was chosen to be half of the detection window length.
For real-world datasets where there is no ground truth drift localization, we follow~\cite{tahmasbi2020driftsurf} and other prior works in reporting average model performance computed across the data stream. We use average accuracy for classification problems and average $R^2$ value for regression problems. In this metric, the practice followed is to retrain the model when a drift is detected, to maintain model performance. We report this metric also for synthetic datasets. 

\label{para:results}
\textbf{Results:} Table \ref{table:det_results} reports our results for drift detection. \name consistently shows the best model performance across all datasets. \name is consistently strong on all drift detection metrics when compared to baseline methods, especially considering it also provides interpretability (discussed later in this section). In particular, \name outperforms the interpretable baselines (\textit{Marginal} and \textit{Conditional}) on all synthetic datasets, and especially on datasets that contain posterior shifts (\textit{Sine} \& \textit{Aug-Mixed}). For real-world datasets, \textit{Marginal} and \textit{Conditional} detect drifts at every new window since they were not designed to detect only model drift, making them ineffective in practice. \cite{kulinski2020feature} made a similar observation. 
\noindent \textbf{Implementation Details.}
The test statistic $\widehat{c}^{k}_{n}(h)$(for a feature $k$) is computed using $\mathcal{Z}_R \sim p(\bm{X},Y),\mathcal{Z}_N \sim q(\bm{X},Y)$ and the model $h$. The model $h\in \mathcal{H}$ is learned by minimizing empirical risk on samples drawn from the distribution $ p(\bm{X},Y)$. To evaluate model risk $h$, we use a held-out set of samples. Therefore, $\mathcal{Z}_R$ samples are kept disjoint from the training samples for $h$ to ensure valid estimation of risk. Given a set of $n$ samples, we train a model on the first $\lfloor nr \rfloor$ samples, and the reference window $\mathcal{Z}_R$ contains the remaining $n- \lfloor nr \rfloor$ samples. $r$ defines the proportion of samples used for training the model to computing the test statistic. $r=0.8$ for all our experiments in this section across all methods, for fairness of comparison. \srinivas{A larger fraction of training samples results in a well trained model. We study the effect of varying $r$ in Sec \ref{ablation}, Table ~\ref{table:ablation_on_r}.} \\
For the bootstrap procedure used in Algorithm \ref{alg:sup}, following \cite{kulinski2020feature}, we use $K=100$ bootstraps and $\alpha=0.05$. For more details on the bootstrap procedure please refer to \cite{kulinski2020feature}. We use default parameters suggested in the respective papers for all baselines. \\ 
As described in Sec \ref{desc_algo}, \name uses a moving window procedure to detect drift on the input data stream. We follow the protocol used by \cite{kulinski2020feature} in this context.
When there is no drift, the new samples window is shifted by $\delta=50$.
For avg model performance, window size $n=1000$ for all datasets and $n=1500$ for measuring precision and recall in synthetic datasets. A larger window size is used for syn datasets considering the availability of data in this setting. This also allows learning better models across the methods when estimating $P_{det}$ \& $R_{det}$. \srinivas{Choosing the appropriate window size is a task dependent issue. If the data stream is prone to many drifts (e.g. if known from domain knowledge), the window size should be small, as a large window size would lead to a delay in drift detection. If the task at hand is known to have drifts rarely, a slightly larger window size could be used as that would improve the quality of the hypothesis testing framework.}\\ 
Base model is a 2-layer neural network for all our experiments in this section. \srinivas{We use a 2-layer neural network since it is the simplest model which performs well on all datasets and allows for quick empirical evaluation. However, \name detect drift for all model classes.} We perform ablation studies in Sec \ref{ablation} to study the impact of different window sizes and model classes on our method's performance. $\delta$ is generally much smaller than window size $n$ to detect drifts close to their true locations. When a drift is detected, a batch of $n$ samples is requested and a new model is trained on the first $\lfloor nr \rfloor$ samples of that batch and reference window is updated with the remaining $n- \lfloor nr \rfloor$ samples. New samples window is shifted by $n$.

\vspace{-12pt}
\subsection{Model Drift Interpretability}
\label{sec:interp}
\name not just detects drifts, but also provides interpretability in terms of input features that are most attributed to the drift. 
In order to study this, we conduct three kinds of studies: (i) We construct new synthetic datasets with known feature attributions and study the precision and recall for the specific features; (ii) We adapt the commonly used occlusion-based metric for interpretability to this setting, and study the change in model performance on datasets studied in Sec \ref{subsec_drift_detection_results}; and (iii) We perform a qualitative case study on the real-world USENET2 dataset to study interpretability in particular. We compare our method against \textit{Marginal} and \textit{Conditional}, which are the interpretable baselines.

\noindent \textbf{Study on Synthetic Datasets:} We construct two synthetic datasets D1 \& D2 with known feature attributions, and report precision and recall of identifying these features in Table \ref{table:feat_loc}. 
D1 is a binary classification dataset on $3$ binary variables $(x_1,x_2,x_3)$ with a single drift. The decision rule pre-drift is $y_{pre} = (x_1 \oplus x_2) \cup x_3$ which post-drift becomes $y_{post} = x_1 \cup x_3$. Here, $y_{pre}$ depends on the pair $(x_1,x_2)$, whereas in $y_{post}$ the pair is not important. Thus, there is a drift due to the change in the relationship between $(x_1,x_2)$ and $y$. Similarly, D2 has 4 binary variables with a single drift caused to change in decision rule from $(x_1 \wedge x_2)$ to $(x_1 \wedge x_3)$. The cause of drift is change in the relation of $(x_2,x_3)$ and y. These datasets have non-linear prediction rules and are challenging for our interpretable baselines.  Results in Table \ref{table:feat_loc} show that \name is able to identify relevant features better than \textit{Marginal} and \textit{Conditional} when tested under non-benign shifts.
\begin{table}[h]
\setlength{\tabcolsep}{3pt} 
\vspace{-7pt}
  \centering
    \caption{\footnotesize Precision ($P$) and Recall ($R$) for feature localization on synthetic datasets with known feature attribution}
    \label{table:feat_loc}
    \vspace{-7pt}
    \scalebox{0.9}{
  \begin{tabular}{cccc|ccc}
    \toprule
    \textbf{Metrics} $\rightarrow$
    & \multicolumn{3}{c|}{\textbf{$P$}} & \multicolumn{3}{c}{\textbf{$R$}} \\
    \textbf{Datasets} $\downarrow$ & \name & Marg & Cond & \name & Marg & Cond\\
    \hline
    D1 & \textbf{1.0} & 0 & 0 & \textbf{1.0} & 0 & 0 \\
    D2 & \textbf{1.0} & \textbf{1.0} & \textbf{1.0} & \textbf{1.0} & \textbf{1.0} & 0.5 \\
    \bottomrule
  \end{tabular}}
  \vspace{-2pt}
\end{table}

\noindent \textbf{Study with Occlusion-based Metric:} 
Following the well-known strategy of occlusion in quantifying interpretability~\cite{pmlr-v97-ancona19a, Zeiler2013VisualizingAU}, we occlude the features identified by a method, impute them with their average value and look at the change in accuracy drop.
Let $\Delta\mathcal{A}([d])= A_{\mathcal{Z}_{R}}^{[d]}(h) - A_{\mathcal{Z}_{N}}^{[d]}(h)$ denote the difference in average accuracy of model $h$ on reference samples $\mathcal{Z_R}$ and new samples $\mathcal{Z_N}$ when all $d$ features are considered. When a drift is detected, we compute the importance of feature subset $S$ as $\sigma(S) = \Delta\mathcal{A}([d]) - \Delta\mathcal{A}([d]\backslash{S})$. One would expect a higher value of $\sigma$ when the most important features are occluded. We define the occlusion metric across a data stream containing $k$ drifts as $\bar{\sigma} = \frac{1}{k}\sum_{i=1}^{k}\sigma(S_i)$, where $S_i$ is the feature subset selected as causing the drift. 
We show results for this study in Table \ref{table:occ_metric}. \name consistently outperforms other interpretable baseline methods.
\begin{table}[h]
\vspace{-5pt}
  \setlength{\tabcolsep}{4pt} 
  \centering
  \caption{\footnotesize Model drift leads to a drop in model accuracy, and features important to drift contribute the most to this drop in accuracy. We report our occlusion metric $\bar{\sigma}$ in percentage. Higher is better. 
  }
  \vspace{-3pt}
    \label{table:occ_metric}
    \scalebox{0.85}{
  \begin{tabular}{lccc}
    \toprule
    \textbf{Datasets} $\downarrow$ \textbf{Methods} $\rightarrow $ & \name & Marginal & Conditional \\
    \midrule
    D1    & \textbf{2.4} & 0.0 & 0.0  \\
    D2    & \textbf{60.0} & \textbf{60.0} & \textbf{60.0}\\
    Sine (balance)    & \textbf{32.0} & 31.1 & 17.0\\
    Sine (imbalance)  & \textbf{25.3} & 20.6 & 17.1\\
    Mixed    & \textbf{0.2} & -14.1 & -11.2\\
    Aug-Mixed        & \textbf{7.6} & -6.0 & 2.0\\
    SEA & \textbf{1.0} & -2.4 & 0.5\\
    SEA-Gradual & \textbf{1.7} & 0.2 & -0.4\\
    Hyperplane & \textbf{0.63} & 0.05 & -0.6\\
    \bottomrule
  \end{tabular}}
\end{table}

\begin{figure}[t]
\small
\centering
    \includegraphics[width=6.8cm, height=3.5cm]{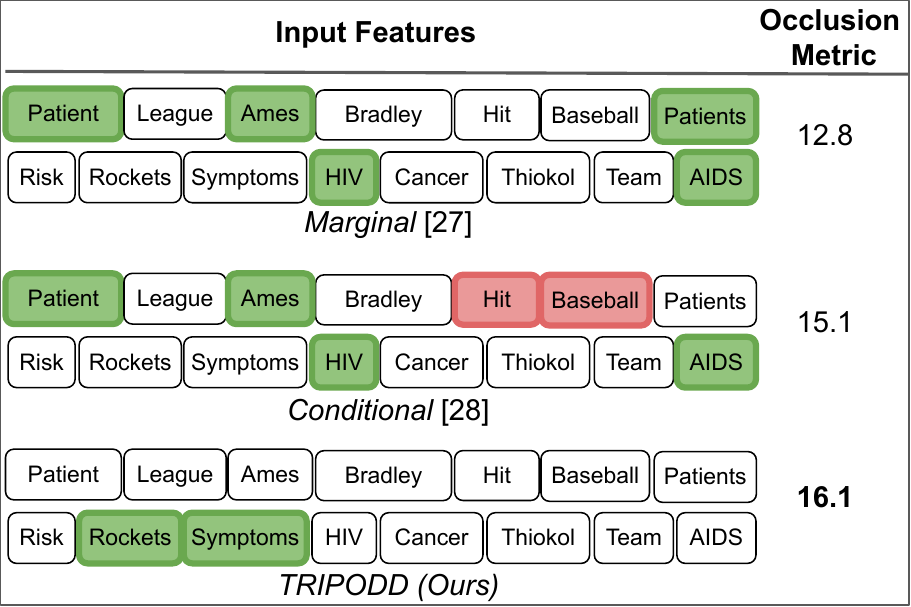} 
\vspace{-8pt}
\caption{\footnotesize\textit{Qualitative Case Study for Drift Interpretability on USENET2 Dataset:} 
Highlighted cells (green or red) are words related to drift w.r.t. each method. Green indicates a correct semantic connection to the drift, red indicates semantically irrelevant words. Our method selects relevant words with a higher sensitivity (as shown on the occlusion metric value). More details in Sec \ref{sec:interp}). \textit{Note:} \textit{Bradley} = baseball team \& \textit{Ames} = NASA center }
\label{fig:use2_res}
\vspace{-2pt}
\end{figure}

\vspace{-8pt}
\noindent \textbf{Qualitative Case Study:} For this study, we use a real-world dataset USENET2 \cite{katakis2008ensemble}, which was obtained by asking people with different interests to label an email message as interesting or not interesting. There are three primary interest groups present in the dataset: \textit{Space}, \textit{Medicine} and \textit{Baseball}. We use a subset of the dataset containing samples from two interest groups - \textit{Space} \& \textit{Medicine}. Thus, this subset of the dataset contains a single drift between samples where people were interested in the \textit{Space} category (size of sample set = 400) vs the samples where people were interested in the \textit{Medicine} category (size of sample set = 400). As we know the cause of the drift in these two sets of samples (change in interest causing a change in posterior distribution), we can study the goodness of interpretations of the drift. We use this dataset since the feature semantics are easy to follow and do not require any background domain knowledge.
Figure \ref{fig:use2_res} shows the result. \name attributes model drift to words (here features) that are semantically related to the user interest categories that are involved in the drift i.e., \textit{medicine} and \textit{space}. Conditional attributes drift to a non-relevant interest category \textit{baseball}. While Marginal also provides correct attributions, the occlusion metric values show higher attribution to the features identified by \name. 
\vspace{-4pt}
\section{Discussion and Analysis}
\label{ablation}

\noindent \textbf{Effect of window size.} We study the effect of window size on drift detection and interpretability performance. 
The black-box drift detectors use adaptive windows or have recommended window sizes, thus we perform sensitivity analysis for drift detection of our method only.
$\delta=50$ for window sizes greater than or equal to 1000, and $\delta=10$ for window sizes less than 1000 (for better sensitivity of drifts). We use a 2-layer neural network as our base model in this study. The results in Table \ref{table:abl_win_det} show that drift detection performance is not affected significantly by a change in window size. However, in a data stream, increasing the window size beyond a certain point can cause an increase in drift localization error (such as in the case of the Electricity dataset for a window size of 1500). We further conduct a similar study for drift interpretability in Table \ref{table:abl_win_int}. The results show our method consistently performs better than the baseline methods.


\begin{table}[!ht]
  \vspace{-5pt}
  \centering
  
  \begin{minipage}{0.47\linewidth}
  \caption{\footnotesize Sensitivity analysis on window size: Average model accuracy of \name for different window sizes.}
    \label{table:abl_win_det}
  \scalebox{0.6}{
  \begin{tabular}{lllll}
    \toprule
    \textbf{Window Size} $\rightarrow $     & 750 & 1000 & 1250 & 1500 \\
    \textbf{Datasets} $\downarrow $ \\
    \midrule
    Hyperplane    & 87.0 & 88.1 & 88.1 & 88.0    \\
    Electricity   & 77.7 & 77.6 & 77.1 & 76.5\\
    Weather    & 77.1  & 77.4 & 77.5 & 77.1  \\
    Powersupply & 72.4  & 72.6 & 73.5 & 72.4  \\
    \bottomrule
    \end{tabular}
    }
  \end{minipage}
  \hspace{5pt}
  \begin{minipage}{0.47\linewidth}
  \vspace{-4pt}
  \caption{\footnotesize Sensitivity analysis on model class: Average model accuracy comparison across different model classes. Here 'k-layer' refers to a k layer neural net. }\label{table:model_class_abl_det}
  \vspace{-10pt}
  \scalebox{0.55}{
  \begin{tabular}{ccc|cc|cc}
  \toprule
    \textbf{Model} $\rightarrow$ & \multicolumn{2}{c|}{2-layer} & \multicolumn{2}{c|}{4-layer} & \multicolumn{2}{c}{6-layer} \\
    \textbf{Methods} $\downarrow$ & Elec & Weat & Elec & Weat & Elec & Weat \\
    \toprule
    \name & \textbf{77.4} & \textbf{77.6} & \textbf{78.7} & \textbf{76.0} & \textbf{76.2} & 74.0 \\
    KSWIN  & 74.7 & 75.6 & 74.2 & 75.0 & 74.3 & \textbf{77.1} \\
    MDDM  & 74.6 & 77.5 & 75.1 & \textbf{76.0} & 76.0 & 76.1 \\
    \bottomrule
  \end{tabular}
    }
  \end{minipage}
\end{table}

\vspace{-24pt}
\begin{table}[!ht]
  \centering
  
  \begin{minipage}{0.47\linewidth}
  \caption{\footnotesize Sensitivity analysis on win size: Occlusion metric comparison across different win sizes on the SEA dataset}
    \label{table:abl_win_int}
  \scalebox{0.6}{
  \begin{tabular}{lllll}
    \toprule
    \textbf{Window Size} $\rightarrow $     & 750 & 1000 & 1250 & 1500 \\
    \textbf{Methods} $\downarrow $ \\
    \midrule
    \name & \textbf{0.10} & \textbf{1.00} & \textbf{0.20} & \textbf{0.10}\\
    Marginal  & -2.10 & -2.40 & -0.27 & \textbf{0.10} \\
    Conditional  & -3.70 & 0.50 & 0.00 & 0.00\\
    \bottomrule
    \end{tabular}
    }
  \end{minipage}
  \hspace{5pt}
  \begin{minipage}{0.47\linewidth}
  \caption{\footnotesize Sensitivity analysis on model class: Occlusion metric comparison across model classes on SEA dataset. Here 'k-layer' refers to a k layer neural net. }\label{table:model_class_abl_occ}
  \vspace{-10pt}
  \scalebox{0.7}{
  \begin{tabular}{cccc}
  \toprule
    \textbf{Methods} $\downarrow$ & 2-layer & 4-layer & 6-layer \\
    \toprule
    \name & \textbf{1.0} & \textbf{0.1} & \textbf{0.3}\\
    Marginal  & -2.4 & -1.62 & -1.69\\
    Conditional  & 0.5 & 0 & 0\\
    \bottomrule
  \end{tabular}
    }
  \end{minipage}
\end{table}

\vspace{10pt}
\noindent \textbf{Effect of Model Class.} Here, we study the drift detection and interpretability performance of \name under different model classes and compare it relevant baselines. We experiment with 3 different model classes: 2-layer (1024 \& 512 neurons), 4-layer (1024, 512, 256 \& 128 neurons) \& 6-layer neural networks (1024, 512, 256, 128, 128 \& 64 neurons) in Table \ref{table:model_class_abl_det}. 
As can be seen from the table, \name performs at par or better than the black-box drift detectors, across model classes, while being interpretable at the same time. We further conduct a similar study for drift interpretability in Table \ref{table:model_class_abl_occ}.
\begin{wraptable}[9]{r}{4cm}
  \vspace{-15pt}
  \centering
  \setlength{\tabcolsep}{3pt}
  \small
  \caption{\footnotesize Average classification accuracy for different values of $r$. Window size = 1000, base model = 2 layer neural network}
  \label{table:ablation_on_r}
  \begin{tabular}{llll}
    \toprule
    Values of $r$ $\rightarrow $     & \hspace{3pt} $0.6$ & \hspace{3pt} $0.7$ & \hspace{3pt} $0.8$\\
    Datasets $\downarrow $ &  &  & \\
    \midrule
    Electricity  & 76.2 & 76.8 & 77.6 \\    
    Weather & 76.1 & 76.9 & 77.4 \\    
    \bottomrule
    \end{tabular}
\end{wraptable}
The results show our method consistently performs better than the baseline methods.
\noindent \textbf{Effect of $\mathbf{r}$:} The ratio that defines the proportion of samples used for training the model to compute the test statistic in the reference window is referred to as $r$. Results reported in Table \ref{table:det_results} were with the $r$ value of $0.8$. Here, in Table \ref{table:ablation_on_r}, we vary the value of $r$ and study its impact on the drift detection performance of \name measured using the metric of average model performance across the data stream. It can be observed from Table \ref{table:ablation_on_r} that the drift detection performance of \name stays approximately the same(within 1-2\%) when $r$ is varied indicating robustness of \name to $r$ value. However, it is useful to note that if $r$ is too high(very close to 1) then there would be very few samples left to calculate the test statistic and the risk estimate would be poor, leading to poor drift detection and interpretation. Similarly, if $r$ is set to a very small value(very close to 0) then the base model would have to be learned on a small set of samples.

\noindent \textbf{Time Complexity.} 
Considering \name takes into account interactions of each feature with all possible subsets of features (similar to Shapley values or MCI scores), it incurs a time overhead to provide useful interpretations. However, the real-world benchmarks datasets used for drift detection have sampling rates in the order of hours (Electricity\cite{harries1999splice} dataset) or even days (Weather\cite{elwell2011incremental} dataset). Our method \name takes $<$ 150 secs for Electricity dataset, and $<$ 280 secs for Weather dataset, to perform the test on a window of size 1000 (both reference and new samples) with a 2-layer neural network. Considering the relative sampling rates of the datasets, our method is pseudo real-time, and is practically relevant \& useful. 


\vspace{-0.4cm}
\section{Conclusions}
\vspace{-4pt}
We propose \name to solve a contemporarily relevant, albeit less-studied problem of model drift detection. We define model drift in terms of input features and come up with a feature-wise hypothesis testing framework for detecting drift in an interpretable manner. We do not observe a trade-off between drift detection performance and interpretability -- \name is interpretable and performs on par with state-of-the-art methods.  
Our method uses only model risk and can be applied to both classification and regression tasks, making it a general-purpose method. We conduct an extensive suite of experiments for drift detection and show that \name provides superior interpretability than existing methods, while performing at par or better than even black-box state-of-the-art methods for drift detection. 
We note that \textit{model drift} can exist even when there is no \textit{feature-sensitive model drift} - for eg. drift due to synergy of a subset of features. Similar to existing feature-interpretable methods in literature \cite{dos2016fast,kulinski2020feature}, we are interested in drifts that can be interpreted in terms of individual features in this work. Extending our method for multivariate model drift detection is a promising future direction. 

\vspace{-5pt}
\section{Appendix}
\noindent\textbf{Proof of theorem 4.3}
\vspace{-8pt}
\footnotesize
\begin{proof}
\label{proof}
First we show that $\widehat{d}^{k}\xrightarrow{p} d^{k}$ asymptotically. Subsequently, we use this fact to find a bound on $|\widehat{c}^{k}_{n} - c^{k}_{n}|$, which we use to derive a lower bound on the test power. 

Let $d^k(h)=\max_{S\subseteq F} | (\mathcal{R}^{S}_{p}(h) - \mathcal{R}^{S\cup \{k\}}_{p}(h)) - 
(\mathcal{R}^{S}_{q}(h) - \mathcal{R}^{S\cup \{k\}}_{q}(h)) | \\  =\max_{S\subseteq F}\Delta(S,k) \implies d^k(h)=\Delta(S^*,k) \text{, where } S^*\text{is the subset which} \\ \text{maximizes } \Delta(S,k)$. Similarly, we can define its sample counterpart as follows: $\hat{d}^k(h)=\Delta_D(S^{**},k)$ , where $S^{**}$ is the subset which maximizes  $\Delta_D(S,k)$. From now on we drop $h$ from $\hat{d}^k(h) \ \ \& \ \ d^k(h)$ for simplicity of notation.
\[\textbf{Convergence in Probability of $\hat{d}^k$ to $d^k$}\]
\footnotesize
\begin{gather*}
\hat{d}{^{k}} -d^{k} =\Delta _{D}\left( S^{*} ,k\right) -\Delta \left( S^{**} ,k\right) \leq \Delta _{D}\left( S^{*} ,k\right) -\Delta \left( S^{*} ,k\right) \\
\leq \max_{S\subseteq F} |\Delta _{D}( S,k) -\Delta ( S,k) |
\end{gather*}
\footnotesize
Similarly we can show that $d^{k} - \hat{d}{^{k}} \leq \max_{S\subseteq F} |\Delta _{D}( S,k) -\Delta ( S,k) |$. Therefore, we can conclude: 
\footnotesize
\begin{equation}
    |d^{k} - \hat{d}{^{k}}| \leq \max_{S\subseteq F} |\Delta _{D}( S,k) -\Delta ( S,k) |
\end{equation}
\footnotesize
Now we find an upper bound of $|\Delta _{D}( S,k) -\Delta ( S,k) |$ with respect to error terms to ultimately upper bound the absolute difference between $d^k$ and $\hat{d}^k$.
\footnotesize
\begin{gather*}
|\Delta _{D}( S,k) -\Delta ( S,k) |=|\ |\underbrace{\left( \hat{\mathcal{R}}_{D_{p}}^{S}( h) -\hat{\mathcal{R}}_{D_{p}}^{S\cup \{k\}}( h)\right)}_{e_{1}} - \\
\underbrace{\left( \hat{\mathcal{R}}_{D_{q}}^{S}( h) -\hat{\mathcal{R}}_{D_{q}}^{S\cup \{k\}}( h)\right)}_{e_{2}} |-
|\underbrace{\left( \mathcal{R}_{p}^{S}( h) - \mathcal{R}_{p}^{S\cup \{k\}}( h)\right)}_{e_{3}} - \\
\underbrace{\left( \mathcal{R}_{q}^{S}( h) -\mathcal{R}_{q}^{S\cup \{k\}}( h)\right)}_{e_{4}} |\ |\\
=|\ |e_{1} -e_{2} |-|e_{3} -e_{4} |\ \leq |e_{1} -e_{3} |+|e_{4} -e_{2} |
\end{gather*}
Substituting back the values,
\begin{gather*}
\ |\hat{d}^{k} -d^{k} |\leq \max_{S\subseteq F}( \ |\hat{\mathcal{R}}_{D_{p}}^{S}( h) -\mathcal{R}_{p}^{S}( h) |+ |\mathcal{R}_{p}^{S\cup \{k\}}( h) -\hat{\mathcal{R}}_{D_{p}}^{S\cup \{k\}}( h) |\\
 + |\hat{\mathcal{R}}_{D_{q}}^{S}( h) -\mathcal{R}_{q}^{S}( h)| +|\mathcal{R}_{q}^{S\cup \{k\}}( h) -\hat{\mathcal{R}}_{D_{q}}^{S\cup \{k\}}( h) |\ )
 \end{gather*}
\begin{gather*}
=\max_{S\subseteq F} |\hat{\mathcal{R}}_{D_{p}}^{S}( h) -\mathcal{R}_{p}^{S}( h) |+\max_{S\subseteq F} |\mathcal{R}_{p}^{S\cup \{k\}}( h) -\hat{\mathcal{R}}_{D_{p}}^{S\cup \{k\}}( h) |
\end{gather*}
\begin{gather*}
+
\max_{S\subseteq F} |\hat{\mathcal{R}}_{D_{q}}^{S}( h) -\mathcal{R}_{q}^{S}( h) |+\max_{S\subseteq F} |\mathcal{R}_{q}^{S\cup \{k\}}( h) -\hat{\mathcal{R}}_{D_{q}}^{S\cup \{k\}}( h) |
\end{gather*}
\footnotesize
Now, we derive the lower bound on the probability of $\widehat{d}^{k}$ being different from $d^k$.
\footnotesize
\begin{gather*}
P\left( |\hat{d}^{k} -d^{k} |\leq \varepsilon \right) \leq P(\max_{S\subseteq F} |\hat{\mathcal{R}}_{D_{p}}^{S}( h) -\mathcal{R}_{p}^{S}( h) |+\\
\max_{S\subseteq F} |\mathcal{R}_{p}^{S\cup \{k\}}( h) -\hat{\mathcal{R}}_{D_{p}}^{S\cup \{k\}}( h) |+
\max_{S\subseteq F} |\hat{\mathcal{R}}_{D_{q}}^{S}( h) -\mathcal{R}_{q}^{S}( h) |
\end{gather*}
\begin{gather*}
+ \max_{S\subseteq F} |\mathcal{R}_{q}^{S\cup \{k\}}( h) -\hat{\mathcal{R}}_{D_{q}}^{S\cup \{k\}}( h) |\ \leq \varepsilon )\\
\text{Using union bound of the following form - } \\ P\left(\sum _{i=1}^{m} x_{i} \geq t\right) \leq \sum _{i=1}^{m} P\left( x_{i}  >\frac{t}{m}\right) \text{ we get},
\end{gather*}
\vspace{-2pt}
\begin{equation}
\label{eqn:bound_on_d}
\begin{split}
    P\left( |\hat{d}^{k} -d^{k} |\leq \varepsilon \right) \geq 1-P\left(\max_{S\subseteq F} |\hat{\mathcal{R}}_{D_{p}}^{S}( h) -\mathcal{R}_{p}^{S}( h) |\geq \frac{\varepsilon }{4}\right) \\
    -\ P\left(\max_{S\subseteq F} |\mathcal{R}_{p}^{S\cup \{k\}}( h) -\hat{\mathcal{R}}_{D_{p}}^{S\cup \{k\}}( h) |\geq \frac{\varepsilon }{4}\right)\\
\ \ \ \ \ \ \ \ \ -\ P\left(\max_{S\subseteq F} |\hat{\mathcal{R}}_{D_{q}}^{S}( h) -\mathcal{R}_{q}^{S}( h) |\geq \frac{\varepsilon }{4}\right) \\
-\ P\left(\max_{S\subseteq F} |\mathcal{R}_{q}^{S\cup \{k\}}( h) -\hat{\mathcal{R}}_{D_{q}}^{S\cup \{k\}}( h) |\geq \frac{\varepsilon }{4}\right)
\end{split}
\end{equation}
\footnotesize
\vspace{-2pt}
We use union bound and Hoeffding's inequality to simplify each term in the RHS in the above equation.
\vspace{-2pt}
\footnotesize
\begin{gather*}
P\left(\max_{S\subseteq F} |\hat{\mathcal{R}}_{D_{p}}^{S}( h) -\mathcal{R}_{p}^{S}( h) |\geq \frac{\varepsilon }{4}\right) \leq 
P\left(\bigcup _{S\subseteq F}\left\{|\hat{\mathcal{R}}_{D_{p}}^{S}( h) -\mathcal{R}_{p}^{S}( h) |\geq \frac{\varepsilon }{4}\right\}\right) \\
 \leq \sum _{S\subseteq F} P\left( |\hat{\mathcal{R}}_{D_{p}}^{S}( h) -\mathcal{R}_{p}^{S}( h) |\geq \frac{\varepsilon }{4}\right)
\leq 2^{|F|+1}\exp\left(\frac{-n\varepsilon ^{2}}{8( M-m)^{2}}\right)
\end{gather*}
\footnotesize
Note: $m \leq |\mathcal{R}_{p}^{S}( h) -\hat{\mathcal{R}}_{D_{p}}^{S}(h)| \leq M \ \ \forall S\subset F$\footnote{This can be achieved with bounded loss functions. We assume this to be a practically reasonable assumption as in practice the loss values do not go out of bounds for models whose training does not diverge.}

We get similar bounds for rest of the terms in RHS of Eq \ref{eqn:bound_on_d},
\footnotesize
\begin{gather*}
P\left(\max_{S\subseteq F} |\mathcal{R}_{p}^{S\cup \{k\}}( h) -\hat{\mathcal{R}}_{D_{p}}^{S\cup \{k\}}( h) |\geq \frac{\varepsilon }{4}\right) \leq
2^{|F|+1}\exp\left(\frac{-n\varepsilon ^{2}}{8( M-m)^{2}}\right)\\
P\left(\max_{S\subseteq F} |\hat{\mathcal{R}}_{D_{q}}^{S}( h) -\mathcal{R}_{q}^{S}( h) |\geq \frac{\varepsilon }{4}\right) \leq
2^{|F|+1}\exp\left(\frac{-n\varepsilon ^{2}}{8( M-m)^{2}}\right)\\
P\left(\max_{S\subseteq F} |\mathcal{R}_{q}^{S\cup \{k\}}( h) -\hat{\mathcal{R}}_{D_{q}}^{S\cup \{k\}}( h) |\geq \frac{\varepsilon }{4}\right) \leq
2^{|F|+1}\exp\left(\frac{-n\varepsilon ^{2}}{8( M-m)^{2}}\right)
\end{gather*}
\footnotesize
Thus, by using the above bounds we can simplify Equation \ref{eqn:bound_on_d} and also show convergence in probability of $\widehat{d}^k$ as follows:
\footnotesize
\begin{equation}
\label{eqn:d_upper_bound}
P\left( |\hat{d}^{k} -d^{k} |\leq \varepsilon \right) \geq 1-2^{|F|+3}\exp\left(\frac{-n\varepsilon ^{2}}{8( M-m)^{2}}\right)
\end{equation}
$\ \ \ \ \ \ \ \ \ \ \ \ \Longrightarrow \lim _{n\rightarrow \infty } P\left( |\hat{d}^{k} -d^{k} | >\varepsilon \right) =0\ \ \forall \varepsilon  >0$
\footnotesize

\vspace{-5pt}
\[\textbf{Convergence of Test Power}\]
From Equation \ref{eqn:d_upper_bound} we can infer the following:
\footnotesize
\begin{equation*}
P\left( |\hat{c}^{k} -c^{k} |\leq \varepsilon \right) \geq 1-2^{|F|+3}\exp\left(\frac{-\varepsilon ^{2}}{8n( M-m)^{2}}\right)
\end{equation*}
\footnotesize
The above bound can also be written in the following form:
\footnotesize
\begin{gather*}
\mathbb{P}\left( c_{n}^{k} -\varepsilon \leq \hat{c}_{n}^{k} \leq c_{n}^{k} +\varepsilon \right) \geq 1-2^{|F|+3}\exp\left(\frac{-\varepsilon ^{2}}{8n( M-m)^{2}}\right)\\
\Longrightarrow \mathbb{P}\left(\hat{c}_{n}^{k} \geq c_{n}^{k} -\varepsilon \right) \geq 1-2^{|F|+3}\exp\left(\frac{-\varepsilon ^{2}}{8n( M-m)^{2}}\right)
\end{gather*}
\footnotesize
We now perform a simple reparameterization by substituting $t = c^{k}_n - \varepsilon$ to get a lower bound on the test power.
\small
\begin{gather*}
\mathbb{P}\left(\hat{c}_{n}^{k} \geq t \right) \geq 1-2^{|F|+3}\exp\left(\frac{-\left( c_{n}^{k} -t \right)^{2}}{8n( M-m)^{2}}\right)=1-2^{|F|+3}\exp\left(\frac{-n\left( d^{k} -\frac{t}{n}\right)^{2}}{8( M-m)^{2}}\right)
\end{gather*}
\footnotesize
Under the alternate hypothesis, it is easy to show that $d^k>0$, therefore we conclude that, $\lim _{n\rightarrow \infty } P\left(\hat{c}^{k} \geq t\right) =1\ \ \forall t >0$
\vspace{-2pt}
\end{proof}

\normalsize
\srinivas{\noindent\textbf{A discussion on Interpretability of \name:} A drift can occur due to a combination of many features, but where no no individual feature flags our hypothesis test. In this event, our method will not detect a drift, and will incur a false-negative. We cannot study the interpretability in this case, since no features are
flagged as causing the drift. However this is a rather special case, and in general for tabular datasets with semantic, uncorrelated, features we expect drift to be ‘caused’ by one or many distinct features. In case of correlated features, our method can be easily extended by grouping the correlated features in a single feature. Our occlusion metric deals with the complementary case: In the
event that a drift detection algorithm flags a drift, how relevant are the detected features? This metric does not consider the cases when the algorithm does not flag a drift. Our algorithm has a high recall for many datasets, indicating that there is no sacrifice in detection performance to ensure interpretability.}

\bibliography{main} 
\bibliographystyle{ACM-Reference-Format}

\end{document}